\documentclass[letterpaper]{article}
\usepackage{iccc}

\usepackage{times}
\usepackage{helvet}
\usepackage{courier}
\usepackage{graphicx}
\usepackage{natbib}
\usepackage{hyperref}
\usepackage[table]{xcolor}
\usepackage{tabularx}
\usepackage{soul}
\definecolor{light-gray}{gray}{0.8}
\sethlcolor{light-gray}
\title{Inspiration through Observation: Demonstrating the Influence of Automatically Generated Text on Creative Writing}
\author{Melissa Roemmele\\
Language Weaver (RWS Group)\\
Los Angeles, CA, USA\\
mroemmele@sdl.com\\
}
\begin{document} 
\maketitle
\begin{abstract}
Getting machines to generate text perceived as creative is a long-pursued goal. A growing body of research directs this goal towards augmenting the creative writing abilities of human authors. In this paper, we pursue this objective by analyzing how observing examples of automatically generated text influences writing. In particular, we examine a task referred to as sentence infilling, which involves transforming a list of words into a complete sentence. We emphasize ``storiability'' as a desirable feature of the resulting sentences, where ``storiable'' sentences are those that suggest a story a reader would be curious to hear about. Both humans and an automated system (based on a neural language model) performed this sentence infilling task. In one setting, people wrote sentences on their own; in a different setting, people observed the sentences produced by the model while writing their own sentences. Readers then assigned storiability preferences to the resulting sentences in a subsequent evaluation. We find that human-authored sentences were judged as more storiable when authors observed the generated examples, and that storiability increased as authors derived more semantic content from the examples. This result gives evidence of an ``inspiration through observation'' paradigm for human-computer collaborative writing, through which human writing can be enhanced by text generation models without directly copying their output.\footnote{All code associated with our model, dataset synthesis, and authoring experiments is available at \href{https://github.com/roemmele/InSentive}{github.com/roemmele/InSentive}. The data resulting from the authoring experiments is also available upon request by contacting the authors.}
\end{abstract}

\section{Introduction}

Creative text generation is a significant focal point at the intersection between computational creativity and natural language processing research. The goal behind much of this research is to understand and simulate human creative writing abilities. There is also increasing interest in using this work to augment human creativity. This objective has become especially visible given recent advancements in systems that can directly interface with human-authored text. 

Many existing creative text generation systems can be applied to facilitate human authoring, even if they are not explicitly presented in this way. The clarity of this use case can largely depend on how the system is evaluated. There is no standard design for such evaluations of benefits to human authoring. Much work uses the convention of comparing generated output to human reference output for a given task, either by comparing the features of the text itself or comparing relative human judgments of it. Success by this standard is based on how well the system simulates human writing. One could theorize that the more a system writes like a human, the more it will be able to help other humans write, but further empirical exploration of this is needed. Alternatively, systems that explicitly aim to support human authoring are often evaluated in the context of interactive applications where authors can elicit generated text. Here, the quality of the model can be evaluated according to authors' interaction with the generated output.

In this paper, we focus on an ``inspiration through observation'' paradigm for human interaction with generated text. In many application settings for text generation, this human interaction is dynamic, with system output changing frequently in direct response to user choices. While discovering the best interaction paradigm is a critical objective of research on authoring support, here we minimize the role of user control over the generated text in order to assess the impact of merely observing the text. Authors see examples of generated text that fulfill a particular authoring objective, and they repeat the same task on their own. We compare human authoring outcomes in the absence and presence of these generated examples. This broad methodology could be applied to probe the ability of any system for aiding authoring, even systems that have not previously been assessed for this use case.

Our exploration of this paradigm focuses on a particular authoring task, \textit{sentence infilling}, and a particular authoring objective, which we term \textit{storiability}. Sentence infilling involves expanding a list of words into a full sentence. In our version of this task, the sentences we elicit can be viewed as story excerpts. The construct of storiability is related to previously discussed ideas such `storiness' by \citet{bailey1999searching}, which pertains to the success of a story from a reader's perspective. We define storiability as the degree to which an excerpt (here, a single sentence) alludes to an appealing story. Even though this is a broad definition, we operationalize it through specific instructions in our experiments. Through our experiments we find that observing automatically generated examples of our sentence infilling task helps people better fulfill the storiability authoring objective. This provides evidence for a general inspiration-through-observation framework by which generation systems can improve human authoring.

\section{Background}

As artificial intelligence has progressed, so has the development of Creativity Support Tools (CSTs). CSTs are digital applications intended to augment human abilities in creative endeavors like visual and performance art, music, and writing (see \cite{frich2019mapping} for a review of several applications). CSTs for writing in particular have been boosted by recent advances in natural language generation, making it possible for systems to interface with any unconstrained human-authored text. This includes figurative language like poetry \citep{kantosalo2019human} and metaphors \citep{gero2019}. Advances in story generation \citep[e.g.][]{fan-etal-2018-hierarchical,martin2021} have been showcased by the increasing development of CSTs that support authoring in the narrative domain. One design pattern for these systems involves authors querying a generation model for a ``suggestion'' that can be integrated into their text \citep{clark2018,khalifa2017deeptingle,manjavacas-etal-2017-synthetic,roemmele2018automated}. This enables analysis of what users choose to do with the generated text (e.g. retaining or deleting it) and how their choices are affected by the features of the text \citep{akoury-etal-2020-storium,roemmele2018linguistic,clark-smith-2021-choose}.

Human-computer interaction studies have compared people's writing with and without the use of AI-based tools, showing that these tools do change how people write. Existing work has examined the effect of word and phrase predictions for content like image captions \citep{arnold2020}, emails \citep{buschek2021}, and movie reviews \citep{bhat-etal-2021-people}. For more open-ended creative writing tasks, most research has focused on optimizing and assessing how much people favor the generated content. What is needed is more experimental comparison of how the use of CSTs changes the authoring outcome as perceived by readers. \cite{mizrahi-etal-2020-coming} recently pursued this for the specific task of creating neologisms (i.e. new words). In their work, people wrote neologisms before and after observing automatically generated examples. The results showed that observing these examples helped people produce better neologisms in terms of their perceived creativity. In this paper, we follow a similar approach to examine the intervening effect of generated examples for the sentence infilling task.

\section{Sentence Infilling}

We focus on the specific task of sentence infilling to evaluate our hypotheses about authoring. Given a sequence of input words (e.g. ``he town rain''), which we refer to as a ``prompt'', the infilling task expands the sequence into a complete sentence by inserting additional words (e.g.``he rode his bike to town in the pouring rain.''). We created a dataset for this task and trained an automated model on it, as detailed below.

\subsection{Overview}

Text infilling, alternatively known as expansion or elaboration, has recently attracted significant attention for multiple types of corpora \citep{donahue-etal-2020-enabling,fedus46657,huang-etal-2020-inset,shen2020blank}. There are different configurations of this task based on the length of the text to be infilled. For stories, some work has focused on inserting sentence-length sequences that connect passages \citep{chandu2020reading,ippolito-etal-2019-unsupervised,mori-etal-2020-finding}. A more constrained version of infilling turns it into a cloze (i.e. fill-in-the-blank) task where infilled segments are single words or short phrases. Our infilling model outputs a single sentence given a sequence of words, but no assumptions are made about the number of words to infill. This design is reflected in existing work applied to creative authoring support \citep{ozbal-etal-2013-brainsup,safovich2020fiction}, but it has yet to be examined how automatically infilled sentences affect human performance of this task.

\subsection{Dataset}\label{dataset_section}

We are not aware of any datasets that mirror the design of our particular infilling task, by which sentences can be generated from any arbitrary sequence of words. However, it is easy to simulate an infilling dataset using existing corpora. Given that the task is framed in the context of storytelling, we obtained 10,000 English-language stories from a variety of genres in the BookCorpus \citep{soskkobayashi2018bookcorpus}. We segmented each story into sentences\footnote{All linguistic processing steps used to derive this dataset, including sentence segmentation, word tokenization, and part-of-speech tagging, were performed with the spaCy library: \href{https://spacy.io/}{spacy.io}}, filtering sentences with less than ten words. To derive pairs of prompts and infilled sentences, we randomly dropped between 60-100\% of words in each sentence. We required that the resulting ablated sentence consist of at least 50\% content words (i.e. nouns, verbs, adjectives), since function words that convey little semantic meaning (i.e. pronouns, prepositions, determiners) are more frequent in text. The ablated sentences became the prompts used as the source inputs to the model, whereas the corresponding original sentences were the target infilled outputs generated by the model. The mean number of words in the prompts and infilled sentences was 4.86 and 19.19, respectively. These pairs were divided into 34,172,128 training instances, 897,473 validation instances, and 894,484 test instances fully held-out during training. 

\subsection{Model Design}

\begin{figure}[h!]
\centering
\includegraphics[width=\linewidth]{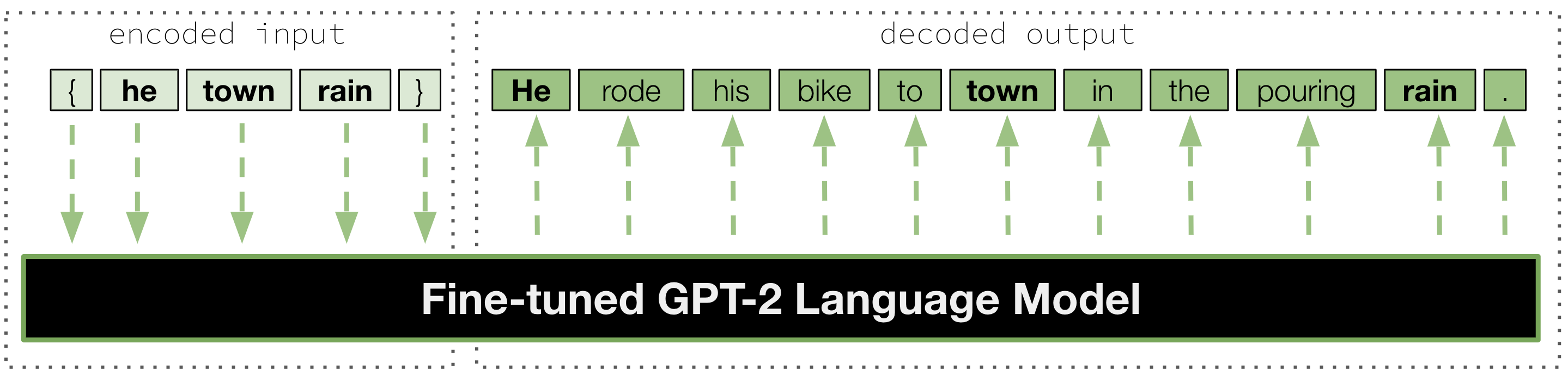}\\
\caption{General architecture of sentence infilling model}
\label{model_architecture}
\end{figure}

Our infilling model\footnote{We used the Texar-PyTorch library for implementation: \href{https://texar-pytorch.readthedocs.io/}{texar-pytorch.readthedocs.io}. Additional hyperparameter settings included: maximum epochs = 100, batch size = 32,
gradient accumulation over 8 steps, validation every 25,000 steps, early stopping after 25 consecutive rounds of no validation improvement, static learning rate = 0.001, maximum gradient norm = 1.0.} is a Transformer language model (LM) \citep{vaswani2017attention}, which is currently a popular architecture for many machine learning approaches to language generation. Figure \ref{model_architecture} broadly illustrates the model. Our scheme for applying this architecture to infilling is closely related to that described in \citet{donahue-etal-2020-enabling}, with one main distinction. Their approach uses designated tokens (i.e. \textsc{[BLANK]}) in the input sequences to indicate the position where text should be infilled in the output. Alternatively, we only represent prompt words in the input, without any explicit signal for where text should be infilled between prompt words. As in the cited work, we initialized the model with weights from pretrained GPT-2 \citep{radford2019language} as a means of embedding general knowledge of English text. GPT-2 has been highlighted for its potential to generate creative text \citep{see-etal-2019-massively,Dathathri2020Plug}. We used the ``small'' version of GPT-2 (117M parameters) and also the corresponding GPT-2 tokenizer to represent all text as subword tokens. We concatenated each prompt and corresponding infilled sentence together as a single token sequence, using designated tokens to signify the start (\textsc{\{\{}) and end (\textsc{\}\}}) of the prompt. To avoid memory errors, we set a limit on the size of the sequences by truncating prompts to the first 25 subword tokens and target sentences to the first 75 tokens. We then fine-tuned the pretrained model for the infilling task by training it on the dataset described above, using the maximum likelihood estimation loss function that is standard for training neural LMs. Our only variation from standard LM training was that we optimized using only the loss for the tokens in the target infilled sentences, and did not compute the loss of the source prompt tokens. This simulates an encoder-decoder scheme which decodes target text from the encoded source input; here the LM functions as both an encoder and decoder, which significantly reduces the number of parameters in the model. We monitored perplexity on the validation items in order to end training when perplexity stopped improving. In inference mode, the model observes a prompt and generates an infilled sentence through a standard LM decoding method. In particular, we autoregressively sample from the LM probability distribution and append the resulting token to the sentence, until the end-of-sequence token (i.e.\textsc{[EOS]}) is generated.

\section{Authoring Experiment}

We next designed a human authoring task that integrates our trained infilling model. To broadly summarize this process detailed in this section: we selected certain prompts from the test partition of our dataset and generated infilled sentences for them. We then elicited human-authored infilled sentences for these same prompts. People produced sentences in two conditions. In the first, they simply wrote sentences for each prompt. In the second, they were shown the sentences generated by our model for the same prompts and wrote new sentences. We explain each of these steps below. 

\begin{figure*}[h!]
\centering
\includegraphics[width=\linewidth]{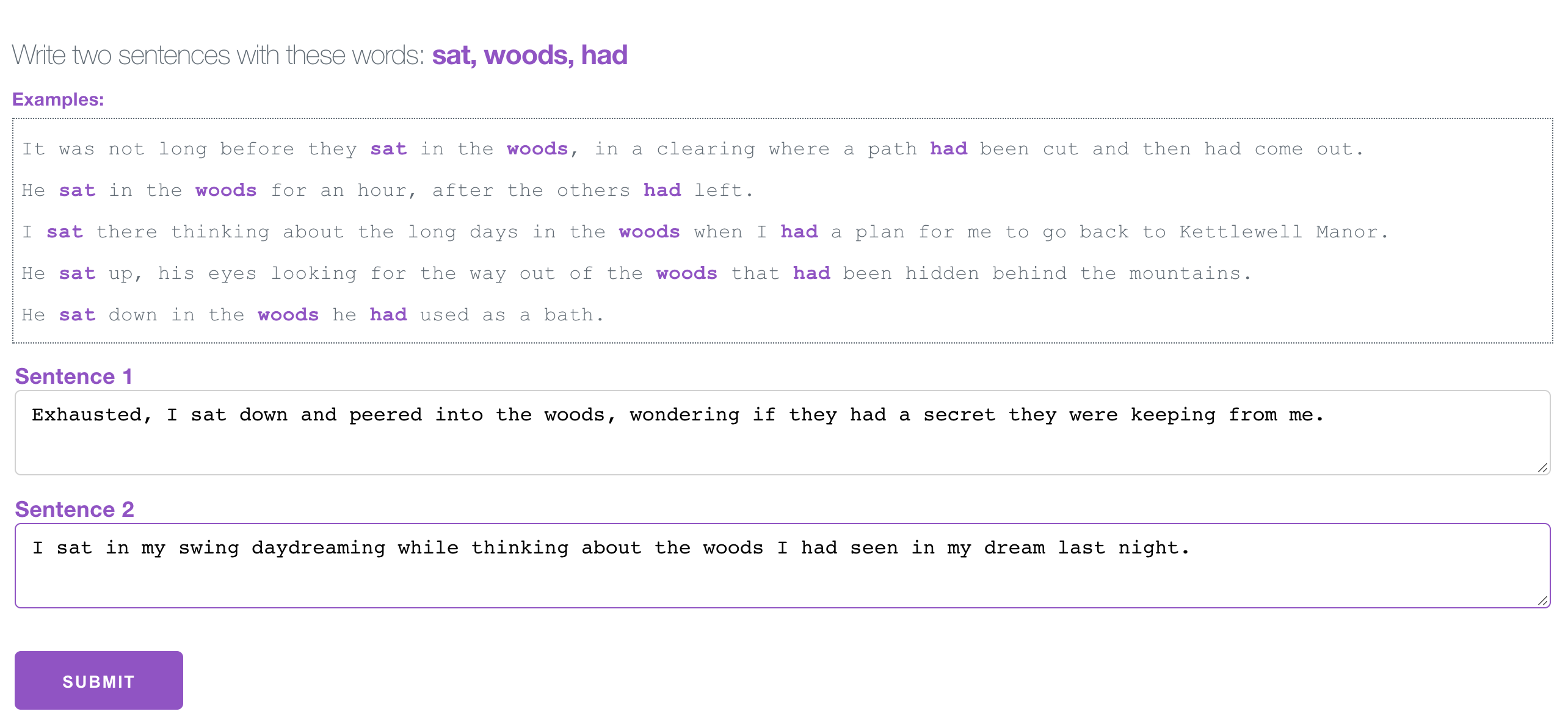}\\
\caption{Screenshot of authoring interface for a single prompt in the \textsc{Post} stage. In the \textsc{Pre} stage, the example sentences are not visible.}
\label{authoring_interface_screenshot}
\end{figure*}

\subsection{Prompt Selection}

We selected prompts from the test set with exactly three words. This particular length value was picked based on intuition. Fewer words approximates unconstrained generation rather than infilling, while more words simulates a constrained fill-in-the-blank task. We excluded prompts derived from dialogue sentences (i.e. those containing quotation marks). Dialogue can pose issues for sentence segmentation (e.g. ``he said.'' may be segmented as a separate sentence from its adjacent quote). We also excluded prompts containing punctuation, numerical digits, named entities\footnote{As with the data creation, this detection was done with spaCy.}, or word tokens not recognized in the DistilBERT (described below) tokenizer vocabulary. Finally, we excluded prompts with more than one function word (e.g. pronouns, prepositions, determiners). By applying these constraints, we expected the prompts to give clear semantic cues for the infilled sentences. The resulting selection consisted of 23,005 prompts. 

Since the process for deriving prompts involved random ablation of full sentences and the position of the ablated words varied, we theorized that even prompts of the same length require different degrees of infilling to yield grammatical sentences. For example, the prompt ``his, body, relax'' already resembles English syntax, and thus it would only take a single infilling word to produce a grammatical sentence (e.g. ``His body could relax''). In contrast, it is possible but harder for native English speakers to find a single infilled word that could transform the prompt ``peculiar, rob, more'' into a grammatical sentence. Accordingly, we expected that the difficulty of the task would vary according to the degree of required infilling for a prompt. We designed an approach for automatically scoring this difficulty. For each selected prompt, we scored the probability of each of its word tokens according to the Masked LM configuration of DistilBERT\footnote{
We used the model interface provided by the HuggingFace transformers library: \href{https://huggingface.co/transformers/}{huggingface.co/transformers}} \citep{sanh2019distilbert}. A Masked LM is well-suited for this measure because it is specifically trained on a fill-in-the-blank task to predict the likelihood of words according to their context. We used the average of the prompt token probabilities to represent the inverse difficulty (i.e. easiness) of a prompt. We theorized that high-probability prompts are easier in infill since they are already probable sequences, whereas low-probability prompts require more infilling to become probable. We assigned the difficulty label ``easy'' to the 10\% highest-probability prompts and the label ``hard'' to 10\% lowest-probability prompts, yielding 2,301 prompts for each difficulty level.

\subsection{Generated Sentences for Prompts}

We then applied the trained model to produce infilled sentences for the selected prompts. We generated five infilled sentences per prompt, using the decoding method of nucleus (top-p) sampling with p = 0.7, based on the parameters recommended by \cite{delucia2020decoding} for generating narrative text. The generated output followed constraints consistent with the human authoring instructions described below. In particular, the generated sentences had to contain all prompt words in the same order as they appeared in the prompt. Prompt words were allowed to be capitalized in the sentence. Sentences had to consist of at least seven word tokens but no more than fifty. We additionally restricted sentences with quotation marks and missing end-of-sentence punctuation (i.e. by requiring the last character to be non-alphanumeric), since this may signify the sequence is not a complete sentence or combines multiple sentences (e.g. quoted dialogue). We filtered sentences with adjacently repeated words (this is a frequently observed issue with neural LMs). Finally, we promoted the diversity of the five sentence outputs for a given prompt by filtering any sentence with 60\% or more of its words already appearing in previously generated sentences for that prompt. All of this criteria was satisfied by continually generating sentences for a prompt until it yielded five acceptable outputs. As a last step that we performed through manual review, we filtered any items where the prompt or generated sentence contained profanity or offensive content. This was done to minimize potential risk of harm to participants in the experiment. The final set consisted of 2,205 easy items and 2,189 hard items. 

\subsection{Human Authoring Task}

\begin{table*}[th!]
\begin{center}
\begin{tabular*}{\linewidth}{p{0.06\linewidth}  l  p{0.17\linewidth}  p{0.175\linewidth}  p{0.4\linewidth}}
\hline
\rowcolor{black} \textcolor{white}{Prompt} & \textcolor{white}{Difficulty} & \textcolor{white}{\textsc{Pre} Sentences} & \textcolor{white}{\textsc{Post} Sentences} & \textcolor{white}{\textsc{Gen} Examples}\\
walking\newline and\newline seeing\newline & easy & \textbf{1.} The little children enjoyed \hl{walking} through the zoo \hl{and} \hl{seeing} all the different animals.\newline \textbf{2.} The boy's favorite activity was \hl{walking} to the marina \hl{and} \hl{seeing} all of the boats in the water. & \textbf{1.} After being released from prison for a crime he didn't commit, the old man was thoroughly enjoying \hl{walking} through the city \hl{and} \hl{seeing} how the world had changed.\newline \textbf{2.} The woman cried when she saw her little girl \hl{walking} \hl{and} \hl{seeing} for the first time after she got her new glasses. & \textbf{1.} She felt the urge to cry, but she kept \hl{walking} \hl{and} \hl{seeing} no sign of it.\newline \textbf{2.} He was \hl{walking} in front of the stove \hl{and} he looked down on the ground \hl{seeing} what was going on.\newline \textbf{3.} We were \hl{walking} in \hl{and} were immediately upon \hl{seeing} what the neighbors had in store.\newline \textbf{4.} She was \hl{walking} with a friend, \hl{and} she just happened to be \hl{seeing} a man, a man, and he was going to kill her.\newline \textbf{5.} She could hear men \hl{walking} up and down the alley, \hl{and} she didn't know what they were doing, but she couldn't deny \hl{seeing} the resemblance.\\
\hline
nose\newline pushed\newline see\newline & hard & \textbf{1.} The sled dogs \hl{nose} was in the air as it \hl{pushed} through the snow to \hl{see} his owner.\newline \textbf{2.} I held my \hl{nose} and \hl{pushed} the stinky garbage can to the curb to \hl{see} if I can catch the garbage man in time. & \textbf{1.} The dog, using his big \hl{nose}, \hl{pushed} the front door open to \hl{see} if his owner was home.\newline \textbf{2.} The boy held his \hl{nose} to stifle a sneeze but the involuntary reflex \hl{pushed} his head forward, watering his eyes and making it hard for him to \hl{see}. & \textbf{1.} The man's \hl{nose} was being \hl{pushed} up and down, and as he moved closer to the screen, the image started to dawn on him, and he was shocked to \hl{see} his father lying on the ground, dying.\newline \textbf{2.} He cleared his throat, the same way he had when he had slapped the back of his head and \hl{nose}, then \hl{pushed} himself away, but he was careful not to let her \hl{see} his anger.\newline \textbf{3.} When he saw his own \hl{nose} in the white sordid mess, he \hl{pushed} off his seat to \hl{see} it for himself. \newline \textbf{4.} He kissed her \hl{nose} and \hl{pushed} the sleeve of her shirt back to \hl{see} what she was thinking.\newline \textbf{5.} A stray \hl{nose}-bleed might be \hl{pushed} up, but I couldn't \hl{see} anything out of place.\\
\end{tabular*}
\caption{Examples of authoring blocks. Each block consists of sentences written by a single author before (\textsc{Pre}) and after (\textsc{Post}) observing the generated (\textsc{Gen}) example sentences.}
\label{example_blocks}
\end{center}
\end{table*}

We then conducted a human authoring task\footnote{ This was implemented as a ReactJS + Flask web application.} utilizing the selected prompts and generated sentences. Participants were instructed that they would be shown a list of three words (the prompt) and would write two unique sentences containing those words. They were presented with some manually written examples of infilled sentences. The instructions emphasized that they should ``try to write sentences that evoke a story someone would be curious to hear'', which activates the construct of storiability that we focus on in this work. The authors' sentences were required to obey the same prompt token order, length, and end-of-sentence punctuation constraints as the model output, which we enforced through the user interface. In the first stage of the task (the \textsc{Pre} stage), each author wrote two sentences for five prompts, which were randomly sampled from the ``easy'' and ``hard'' categories. In the second stage (the \textsc{Post} stage), authors were again shown the same five prompts and wrote an additional two unique sentences for each. This time, the five generated sentences were shown to them as examples they could reference while writing. Their sentences were required to be different from the examples. Figure \ref{authoring_interface_screenshot} shows an example screenshot of the interface for this exercise.

The presence of the generated examples was the only variable that differed between the two stages. In both stages, after submitting the sentences for a single prompt, participants were shown generated text passages (described as ``stories'') that each began with the sentences they wrote. These passages were generated by the original pretrained GPT-2\footnote{Using the model interface provided by HuggingFace transformers; generated using nucleus sampling with p = 0.7} (not the infilling model). Passages had a maximum length of 75 words, and only the first $k$ complete sentences within this limit were displayed. The instructions informed authors that writing more interesting sentences would yield more interesting stories. However, this component of the task was not an experimental variable, since it was not varied between the two stages. This feedback was simply intended to incentivize authors to write more storiable sentences. 

We recruited participants for this task through Amazon Mechnical Turk\footnote{\href{https://www.mturk.com/}{mturk.com}} (AMT), a crowdsourcing platform. 23 authors from majority native English-speaking countries were each paid \$10 based on an estimated completion time of 45 minutes to 1 hour. The result was a dataset of \textit{authoring blocks}, with each block consisting of a prompt shown to an author, their two sentences written before observing the generated examples (\textsc{Pre}), their two sentences written after the observing the generated examples (\textsc{Post}), and the five generated examples they saw (\textsc{Gen}). Examples of authoring blocks are shown in Table \ref{example_blocks}. With each author responding to five unique prompts, this yielded 115 blocks. We filtered six blocks where at least one sentence response (\textsc{Pre} or \textsc{Post}) was revealed to actually consist of multiple sentences (since this wasn't straightforward to check through the interface during the task). This ultimately resulted in a set of 109 blocks to be used for evaluation, 53 for easy 
prompts and 56 for hard prompts.

\section{Evaluation of Authoring Experiment}

\begin{table*}[th!]
\begin{center}
\begin{tabularx}{\linewidth}{  p{0.075\linewidth}  p{0.07\linewidth}  X  X  X}
\hline
\rowcolor{black} \textcolor{white}{Prompt} & \textcolor{white}{Difficulty} & \textcolor{white}{\textsc{Pre} Sentence} & \textcolor{white}{\textsc{Post} Sentence} & \textcolor{white}{\textsc{Gen} Sentence}\\
felt\newline meet\newline again & easy & \textbf{Jenna \hl{felt} a spooky sense of deja vu and felt that she was about to \hl{meet} a familiar stranger yet \hl{again}.} & Bonnie \hl{felt} a syrupy sentimentality and nostalgia and wanted to \hl{meet} Raphael \hl{again}. & I \hl{felt} so relieved to \hl{meet} you \hl{again}.\\
\hline
regard\newline sorts\newline prevent & hard & \textbf{He had no \hl{regard} for his own safety, a maverick of \hl{sorts}, which did nothing to help \hl{prevent} him from oft getting injured.} & In \hl{regard} to the message, there were all \hl{sorts} of interpretations that could be made, so she asked for clarification to \hl{prevent} misunderstandings. & A lower \hl{regard} may come to any type of treatment that may result in a delay of \hl{sorts} in order to \hl{prevent} future evidence of therapy.\\
\hline
servants\newline early\newline life & easy & It's sad when people have \hl{servants} that have to wake up \hl{early} and do everything for someone else without having a \hl{life} of their own. & \textbf{They became \hl{servants} at a very \hl{early} age after having a difficult \hl{life} and losing their parents.} & But, yes, there were two excellent \hl{servants} from a very \hl{early} age in the village, who could carry the \hl{life} of an even younger man.\\
\hline
hoping\newline questions\newline few & hard & I was \hl{hoping} I could find the answer to my homework \hl{questions}, and after a \hl{few} minutes I found them by doing a simple Google search. & \textbf{She pored her thoughts, fears, and dreams into her diary, \hl{hoping} that by writing them down, she could answer the vexing \hl{questions} of life that \hl{few} people ever really understood.} & They were \hl{hoping} to avoid answering any \hl{questions} for a \hl{few} days.\\
\hline
quickly\newline and\newline joined & easy & There was a bird that \hl{quickly} fell from the sky \hl{and} \hl{joined} with the ground. & The car \hl{quickly} entered the lane \hl{and} \hl{joined} with the traffic. & \textbf{The nurse \hl{quickly} packed up the case \hl{and} \hl{joined} him.}\\
\hline
arms\newline awkwardly\newline car & hard & The \hl{arms} hung \hl{awkwardly} out the window of the \hl{car}. & His \hl{arms} flung \hl{awkwardly} as the police slammed him up against the \hl{car} to cuff him. & \textbf{Sue wrapped her \hl{arms} around his neck, pulled him \hl{awkwardly} out of the \hl{car}, and then pushed him down the long, steep driveway.}\\
\hline
\end{tabularx}
\caption{Examples of judgment groups. The bolded sentence in each group was selected by both raters as the most storiable.}
\label{example_judgment_groups}
\end{center}
\end{table*}

In line with the objective of the authoring task, we conducted a judgment task to evaluate readers' perceived storiability of the sentences in the authoring blocks. This resembles story generation evaluations where people are asked which one of a set of stories they most prefer reading  \citep[e.g.][]{fan-etal-2018-hierarchical}. For each of the 109 blocks, we gathered all unique combinations of the two \textsc{Pre} sentences, two \textsc{Post} sentences, and the first two of the observed \textsc{Gen} examples in that block, yielding 872 \textit{judgment groups} ($109*2*2*2=872$). Thus, each judgment group consisted of a \textsc{Pre}, \textsc{Post}, and \textsc{Gen} sentence aligned to the same prompt and author. We designed a questionnaire targeting the relative storiability of the sentences in each group. Raters were instructed to ``imagine that each sentence [in the judgment group] is an excerpt from a story and pick the one that makes you most want to read that story''. Only the sentence text itself was shown, and the sentences in each group were randomly ordered. We recruited 16 participants from majority native English-speaking countries through AMT to rate subsets of 55-56 judgment groups, with each paid \$5 for an estimated completion time of 25-30 minutes. There were two raters for each subset, yielding a total of 1,744 responses (848 for authoring blocks with easy prompts and 896 for hard). Examples of judgment groups are shown in Table \ref{example_judgment_groups}. In these examples the bolded sentence was picked by both of its raters as the most storiable among the group.

For the results described below, we discuss judgments in terms of storiability preferences. In particular, each response is a single data point where the most storiable sentence selected by the rater was labeled as ``Preferred'' and the other sentences in the judgment group were labeled as ``Not Preferred''. All data points have equal weight in the analyses.

\subsection{Results}

\subsubsection{Human versus Generated Storiability}

Table \ref{human_vs_ai} shows the normalized distribution of storiability preferences across the \textsc{Pre}, \textsc{Post}, and \textsc{Gen} sentences, along with their raw number of ``Preferred'' votes. Note that if preferences were randomly distributed across these three sets, each would approximate 0.33 (one-third) of the distribution. The numbers show that people notably preferred human-authored sentences (both \textsc{Pre} and \textsc{Post}) to \textsc{Gen} sentences (statistically significant at $p < 0.05$)\footnote{Statistical significance for all analyses was determined by two-sample Monte Carlo permutation tests.}. 

In contrast with human authoring, the infilling model did not receive any explicit instructions about the storiability authoring objective. The model was simply trained to generate sentences that appeared in stories. We can guess that the training sentences observed by the model are not all equally likely to be perceived as storiable. It is possible that this is why raters favored human-authored sentences over the generated ones. However, even generated text designed to mimic human writing objectives often does not meet this standard \citep[e.g][]{lin2020commongen}, so the difference in preferences is not simple to interpret. The focus of this particular paper is not on comparing the relative quality of human and generated text, but on whether generated text can alter the quality of human writing. Thus, the rest of our analyses concentrate on this question.

\begin{table}[th!]
\begin{center}
\begin{tabular}{  c  c  c  }
\hline
\textbf{Preferred \textsc{Pre}} & \textbf{Preferred \textsc{Post}} &
\textbf{Preferred \textsc{Gen}}\\
0.356 (621) & 0.365 (636) & 0.279 (487) \\
\hline
\end{tabular}
\caption{Distribution of storiability preferences}
\label{human_vs_ai}
\end{center}
\end{table}

\subsubsection{Prompt Difficulty}

Table \ref{prompt_difficulty} shows the effect of difficulty on the number of infilling words people used to connect the prompt words. The human-authored sentences for the hard prompts had significantly more infilled words between prompt words compared with easy prompts ($p < 0.05$). This validates the expected difference between these conditions, suggesting that hard prompts required more authoring effort.

\begin{table}[th!]
\begin{center}
\begin{tabular}{  c  c  }
\hline
\textbf{Difficulty} & \textbf{Infilled Words}\\
easy & 3.035\\
hard & 4.317\\
\hline
\end{tabular}
\caption{Mean number of words between prompt words in human-authored sentences according to difficulty}
\label{prompt_difficulty}
\end{center}
\end{table}

\subsubsection{Prompt Difficulty and Storiability}\label{difficulty_storiability_section}

Table \ref{prompt_difficulty_storiability} shows the distribution of preferences for \textsc{Pre} and \textsc{Post} sentences grouped by difficulty level. We found that \textsc{Post} sentences had higher storiability than the \textsc{Pre} sentences, but only for hard prompts ($p < 0.05$). Thus, people were more likely to write storiable sentences for these prompts after observing the \textsc{Gen} examples. The result for easy prompts showed a tendency towards the reverse pattern, but the difference in this case was not statistically significant.  Based on this result, we focus our subsequent analyses on the items associated with hard prompts. We return to some discussion of this interaction effect regarding difficulty in the next section.

\begin{table}[th!]
\begin{center}
\begin{tabular}{  c  c  c }
\hline
\textbf{Difficulty} & \textbf{Preferred \textsc{Pre}} & \textbf{Preferred \textsc{Post}}\\
easy & 0.384 & 0.354\\
hard & 0.329 & 0.375\\
\hline
\end{tabular}
\caption{Distribution of storiability preferences for human-authored sentences by difficulty}
\label{prompt_difficulty_storiability}
\end{center}
\end{table}

\subsubsection{Influence of Generated Examples}

The higher preference for the \textsc{Post} sentences suggests that observing the \textsc{Gen} examples had some impact on authors. One could consider other interpretations: for example, maybe authors were simply better at the task in the \textsc{Post} stage after a round of practice in the \textsc{Pre} stage. To investigate this, we first determined whether any influence of the \textsc{Gen} examples could be quantitatively detected in the \textsc{Post} sentences. There are many different features that could be used to quantify this influence. Here we focused on whether authors incorporated semantic content from the examples they observed. We assessed this using a quantitative measure of semantic similarity between sentences based on vector representations given by a pretrained language model. Intuitively, pretrained LMs are expected to produce similar vector representations for sentences with a similar meaning. This representation should transcend the lexical level, so that even sentences with few words in common can have a high similarity score if their respective words in context are synonymous. We computed semantic similarity between the \textsc{Pre} and \textsc{Gen} sentences, and then separately between the \textsc{Post} and \textsc{Gen} sentences. Since the \textsc{Gen} examples were not shown in the \textsc{Pre} condition and thus could have no influence on the \textsc{Pre} sentences, any significant difference in this measure between the \textsc{Pre} and \textsc{Post} sentences can be attributed to authors observing the \textsc{Gen} examples.

We computed the cosine vector similarity between sentences encoded with the DistilBERT\footnote{The same core model used for computing probability scores to determine prompt difficulty, as described earlier. Here, we use the raw hidden states of the model for feature representation instead of the Masked LM probability outputs.} LM. For a given prompt, the similarity score for a human-authored sentence $h$ is its maximum similarity over all \textsc{Gen} examples $gs$ for that prompt, i.e. $score(h, gs) =\max_{g \in gs} sim(h, g)$. We select the maximum because there may be one \textsc{Gen} example in particular that most influences a given sentence.  

Table \ref{similarity_generated_examples} shows the mean of this similarity measure for the \textsc{Pre} and \textsc{Post} sentences. \textsc{Post} sentences had higher similarity to \textsc{Gen} sentences ($p < 0.05$), confirming that the \textsc{Gen} examples had semantic influence on the authors' writing.

\begin{table}[th!]
\begin{center}
\begin{tabular}{  c  c }
\hline
\textbf{Condition} & \textbf{Similarity}\\
\textsc{Pre} & 0.921\\
\textsc{Post} & 0.923\\
\hline
\end{tabular}
\caption{Similarity between human and generated sentences before (\textsc{Pre}) and after (\textsc{Post}) observation of \textsc{Gen} examples}
\label{similarity_generated_examples}
\end{center}
\end{table}

\subsubsection{Influence and Storiability}

After verifying that the difference between the \textsc{Pre} and \textsc{Post} conditions can be attributed to semantic influence from the \textsc{Gen} examples, we examined whether this influence was related to the higher storiability of the \textsc{Post} sentences. Table \ref{semantic_influence_storiability} demonstrates that sentences preferred as more storiable were also more semantically influenced by the \textsc{Gen} examples, as indicated by the higher similarity scores for the Preferred sentences ($p < 0.05$). Thus, by incorporating some degree of content from the \textsc{Gen} examples, people tended to better fulfill the authoring objective. Table \ref{influence_examples} gives some examples of judgment groups where semantic influence can be qualitatively observed in the \textsc{Post} sentence. The \textsc{Gen} example with the most influence is shown (i.e. the one most similar to the \textsc{Post} sentence), and we comment on the subjective evidence of their similarity. These results encourage future opportunities for explaining the exact mechanism underlying semantic influence. We discuss this further in the next section.

\begin{table}[th!]
\begin{center}
\begin{tabular}{  c  c }
\hline
\textbf{Judgment} & \textbf{Similarity}\\
Not Preferred & 0.922\\
Preferred & 0.925\\
\hline
\end{tabular}
\caption{Similarity between \textsc{Post} and \textsc{Gen} sentences (i.e. degree of semantic influence) according to storiability preferences}
\label{semantic_influence_storiability}
\end{center}
\end{table}

\begin{table*}[th!]
\begin{center}
\begin{tabularx}{\linewidth}{ p{.08\linewidth} X X X p{.125\linewidth}}
\hline
\rowcolor{black} \textcolor{white}{Prompt} & \textcolor{white}{\textsc{Pre} Sentence} & \textcolor{white}{\textsc{Post} Sentence} & \textcolor{white}{Influential \textsc{Gen} Example} & \textcolor{white}{Description}\\
shoulders\newline waves\newline color & My \hl{shoulders} were aching but I was set on diving through the \hl{waves}, the \hl{color} of the water getting deeper the further out I went. & Her new hair cut had the length to the \hl{shoulders}, with \hl{waves} of a bright pink \hl{color} all the way down. & His hair was cropped short, flowing down his \hl{shoulders}, but there were \hl{waves} of the same \hl{color}. & Connected prompt words via semantic category of hair\\
\hline
there\newline die\newline capacity & The bouncer thought \hl{there} was a chance people might \hl{die} if there was a fire because the club was way over its \hl{capacity}. & \hl{There} is no chance you're not going to \hl{die}, so you have to come to terms with that in some \hl{capacity}. & It seems, that \hl{there} is a good chance that I will \hl{die} in my \hl{capacity} to forgive and to get on with my life. & Used less literal sense of word ``capacity''\\
\hline
meant\newline said\newline store & The child yelled at her sister not understanding what she \hl{meant} when she \hl{said} to her that she wanted some comics from the \hl{store}. & It \hl{meant} a lot to me when she \hl{said} she was going to the toy \hl{store} to get me a game. & It \hl{meant} a lot to me, because I'd \hl{said} I'd drop by the \hl{store}. & Used phrase ``it meant a lot to me''\\
\hline
spent\newline wind\newline him & After his run he stood by the beach, \hl{spent}, as the \hl{wind} whipped by \hl{him}. & She \hl{spent} the day by the water, the \hl{wind} whipping her hair, aching for \hl{him}. & She \hl{spent} the rest of the day in the saddle, keeping the \hl{wind} from blowing through her hair and reminding her of her promise to get \hl{him} a hot bath. & Used expanded form of phrase ``spent the day'' (``spent the rest of the day'')\\
\hline
peculiar\newline rob\newline more & I have a \hl{peculiar} friend named \hl{Rob} who always wants \hl{more} excitement. & I felt it was very \hl{peculiar} that after talking to \hl{Rob} for only about an hour, I wanted to know \hl{more} about him. & They felt a \hl{peculiar} attraction to \hl{Rob}, but couldn't afford to spend much \hl{more} time together. & Referred to curiosity about Rob\\
\hline
\end{tabularx}
\caption{Examples of \textsc{Post} sentences demonstrating semantic influence, with subjective descriptions of how influence is seen. For reference, the \textsc{Pre} sentence without semantic influence is also shown.}
\label{influence_examples}
\end{center}
\end{table*}

\section{Discussion} 

Observing automatically generated examples of sentence infilling influenced authors to better perform this infilling task on their own. Even though this is a contrived exercise different from conventional forms of creative writing, it still calls upon the same linguistic creativity. A related task is reflected in the real world through popular word games where people produce sentences given word constraints and players rate the interpretability and creativity of the resulting sentences \citep[e.g.][]{cooper}. The task is also applicable to CSTs for writing: for example, a writer might want to brainstorm about potential connections between words they already have in mind, which could be facilitated by a model related to infilling. In contrast to other research on CSTs, this paper focuses less on the interactive capabilities of such systems, like enabling author control over generated output, but our findings are still relevant to interactive applications.

We chose to emphasize the authoring objective of storiability because of our focus on AI-augmented story writing. Storiability is not a one-size-fits-all metric for this research. The quality of a story can be judged on multiple dimensions that are often not consistently defined across different studies, as discussed in \cite{celikyilmaz2020evaluation}.  Evaluations tend to target both the sensibility of stories (e.g. grammaticality,  coherence, plausibility) and their more ``creative'' aspects (e.g. interestingness, suspensefulness, humorousness). The notion of storiability is more related to the latter group, but does not preclude other dimensions. For example, if a sentence contains grammatical errors, a person may not prefer to read the story associated with that sentence. By operationalizing storiability according to a specific question (``which sentence makes you want to read more?''), we tried to elicit judgments that encompass many ways this objective can be achieved. Future research can examine more specific formulations of this question.

An intriguing finding was the difference in outcomes according to prompt difficulty, such that only sentences for hard prompts displayed more storiability as an effect of observing generated text, with no such pattern for easy items. This points to a broad direction for future work: to examine how the demands of the writing task itself affect authors' interaction with an automated model. For instance, authors' engagement with writing assistance tools varies at different times during a single writing session, as discussed in \cite{huang2020}. This may be due to some parts of the text being harder to write than others, as hinted by the mediating effect of difficulty in our results. Interestingly, a follow-up analysis showed there were no significant differences in \textsc{Post} similarity to \textsc{Gen} examples based on difficulty, meaning that the \textsc{Gen} sentences for easy prompts had just as much semantic influence as those for hard prompts. Thus, this influence was somehow not as helpful in promoting storiability in the easy case. One possibility is that authors were already good at producing storiable sentences for easy prompts in the \textsc{Pre} stage, so even when they were influenced by the \textsc{Gen} examples, this influence did not additionally benefit the \textsc{Post} sentences. The hard prompts may have been more challenging, giving the \textsc{Gen} examples a larger opportunity to enhance the \textsc{Post} sentences in this case. Because our evaluation did not include pairwise comparisons between sentences for easy and hard prompts, it will require further research to better understand this finding. 

Our analysis of semantic influence confirms authors derived certain content from the observed examples. More investigation is needed to understand what type of content was most influential. Authors may have extracted specific words and phrases, as indicated by some of the examples in 
Table \ref{influence_examples}, but they did not simply copy or mimic the examples at large; if they had, there would not be a significant difference in storiability between the \textsc{Post} and \textsc{Gen} sentences as reported in Table \ref{human_vs_ai}. One thought is that authors utilized an idea conveyed by a \textsc{Gen} sentence, but reformulated the sentence to repair inadequacies such as ill-formed, awkward, or vague wording. It is also possible that the \textsc{Gen} examples revealed a semantic dimension by which the prompt words were related, one that authors did not initially consider in the \textsc{Pre} condition. The first example in Table \ref{influence_examples} might convey this: the \textsc{Gen} example connects the prompt words ``shoulders'', ``waves'', and ``color'' through the conceptual dimension of ``hair''. Perhaps the example triggered the author to recall this particular concept unifying the prompt words, and they emulated it in the \textsc{Post} sentence. One targeted metric for examining influence could focus specifically on modeling this activation of ``latent'' concepts. Our work quantified influence according to a single measure, but future work could attempt to narrow down the influence of specific linguistic features such as syntactic style (e.g. relative proportion of nouns, verbs, prepositions, etc.), emotional tone (e.g. joyful, sorrowful, fearful), and narrative perspective (e.g. references to pronouns and proper nouns). Existing work has addressed this by examining the strategies authors develop for eliciting precise types of content from generation models;  for example, by triggering the model at certain syntactic positions in a sentence \citep{calderwood2020novelists}. We can use these analyses to guide future systems towards producing content authors find most helpful. 

\section{Conclusion}

In this paper, we explore the question of how automatically generated text can influence human creative writing. We specifically assessed this question through the authoring task and objective of sentence infilling and storiability, respectively. In accordance with a proposed inspiration-through-observation paradigm by which automated models provide helpful examples of how to fulfill the task, we found that observing generated sentences enhanced reader-judged appeal of human-authored sentences. Our results provide empirical evidence that automated models can intervene in the writing process without necessarily replacing human effort. This invites further exploration of this paradigm for other authoring tasks and objectives. The outcome has the potential to transcend the standard of both human and computer authoring when each function independently.








\bibliographystyle{iccc}
\bibliography{iccc}

\end{document}